%% file: main.tex
\documentclass{article} %

\PassOptionsToPackage{numbers,compress}{natbib}
\usepackage[final]{style/neurips_2026}
\IfFileExists{style/neurips_2026.sty}{}{}

\makeatletter
\renewcommand{\@notice}{}
\makeatother

\usepackage{booktabs}
\usepackage{amsmath,amssymb}
\usepackage{amsthm}
\usepackage[normalem]{ulem}

\usepackage{graphicx}
\usepackage{array}

\usepackage{pgfplots}
\pgfplotsset{compat=1.18}
\usetikzlibrary{arrows.meta}
\usepackage{microtype}
\usepackage{xcolor}   %
\usepackage{comment}  %
\usepackage{xstring}  %
\usepackage[hidelinks]{hyperref}
\usepackage{url}
\usepackage[inline]{enumitem}

\newif\ifdraft
\drafttrue
\IfFileExists{paper/nodraft.tex}{\draftfalse}{}  %
\ifdraft
  \newcommand{\todo}[2][]{%
    \IfSubStr{#1}{Comment}%
      {\textcolor{violet}{[\textbf{\textsc{#1}:} #2]}}%
      {\textcolor{red}{[\textbf{TODO}\if\relax\detokenize{#1}\relax\else~\textsc{#1}\fi\textbf{:} #2]}}%
  }
  \newcommand{\pending}[1]{\textcolor{blue!60!black}{[\emph{pending: #1}]}}
  
\else
  \newcommand{\todo}[2][]{}
  \newcommand{\pending}[1]{}
  \excludecomment{notes}  %
\fi

\newcommand{\vocab}{\mathcal{V}}           %
\newcommand{\XX}{\mathcal{X}}              %
\newcommand{\rate}{\lambda}                %
\newcommand{\tot}{\Lambda}                 %
\newcommand{\jump}{J}                      %
\newcommand{\val}{q}                       %
\newcommand{\gen}{\mathcal{L}}             %

\title{When Edit Flows are Edit Jumps:\\
replicating Edit Flows and EvoFlows}

\newif\iflatexml
\IfFileExists{latexml.sty}{\usepackage{latexml}}{}  %
\iflatexml
  \author{%
    Gabriel B\'en\'edict~~~~Melanie Buechler~~~~Gerard Riera-Solà~~~~%
    Chlo\'e de Ancos~~~~Yves Gaetan Nana Teukam~~~~Moritz Freidank \\
    Visium, Switzerland \\
    \texttt{\{gabriel.benedict,\ melanie.buechler,\ gerard.riera,\ chloe.deancos,\ yves.nana,\ moritz.freidank\}@visium.com}%
  }
\else
  \author{
    \textbf{Gabriel B\'en\'edict \quad
    Melanie Buechler \quad
    Gerard Riera-Solà \quad
    Chlo\'e de Ancos} \\
    \textbf{Yves Gaetan Nana Teukam \quad
    Moritz Freidank} \\[4pt]
    Visium, Switzerland \\
    \texttt{\{gabriel.benedict,\ melanie.buechler,\ gerard.riera,} \\
    \texttt{chloe.deancos,\ yves.nana,\ moritz.freidank\}@visium.com}
  }
\fi

\begin{document}

\hypersetup{pdftitle={When Edit Flows are Edit Jumps}, pdfauthor={Gabriel B\'en\'edict et al.}}

\maketitle

\begin{abstract}
\input{sections/abstract}
\end{abstract}

\input{sections/introduction}
\input{sections/background}        %
\input{sections/setup}             %
\input{sections/results}           %
\input{sections/discussion}       %
\input{sections/related}

\input{sections/conclusions}      %

\newpage
\input{sections/ai_statement}
\bibliographystyle{plainnat}
\bibliography{refs}

\newpage
\appendix
\input{sections/appendix}

\end{document}

%% file: sections/abstract.tex
Antibody lead optimization calls for a small, bounded set of edits to an existing candidate: substitutions, but also insertions and deletions. Edit-based generative models are the only ones that allocate such an edit budget without fixing the edit positions, the edit count, or the output length in advance. However, the existing approaches Edit Flows and EvoFlows did not release code or complete training specifications. 
Here, we show that both methods follow the same underlying process — edits firing one at a time, at learned rates, in continuous time — the pure-jump case of generator matching over finite sequences. With \textit{EditJumps} we introduce the first open implementation of this framework, with a single generalist antibody editor trained on 1.66M Observed Antibody Space homolog pairs to propose homolog-like variants of a seed sequence, editing unseen leads zero-shot, without the per-family retraining original approaches require. Replicating this system from scratch exposes why open code is essential for generative biology: reconciling published edit distributions required reverse-engineering an undocumented rate-scaling hyperparameter that dictates realized mutation counts. Moreover, we show that published evaluation metrics are highly sensitive to reference sample size, frequently flipping method rankings. We release our full codebase, automated test suite, and configurations at: \url{https://github.com/VisiumCH/editjumps}

%% file: sections/introduction.tex
\section{Introduction}
\label{sec:intro}

Lead optimization of a protein candidate aims to explore the local sequence landscape around a starting molecule by introducing small controlled edits that balance exploratory novelty with evolutionary naturalness. Doing this by rational design is difficult: mutations interact non-linearly across the sequence, so the effect of any one change is difficult to anticipate and the useful edits cannot be enumerated by inspection. A model for this task must therefore be able to place a small number of edits itself, by deciding where and what to change, while allowing the sequence length to vary as it would in natural homologs.

Standard generative architectures rarely provide this full combination. Masked protein language models such as ESM-2 \citep{esm2} resample residues at given masked positions and return a sequence of the input length, so the positions are an input and no length change is possible. Antibody infilling models such as IgLM \citep{iglm} are autoregressive: the span boundaries are supplied by the user, and the span is regenerated left-to-right, so the model can rewrite it entirely and the number of edits is not controlled. Discrete diffusion models such as EvoDiff \citep{evodiff} need no per-position choice unconditionally, but draw the length before decoding and require a mask for conditional generation. Edit-based models in natural language processing \citep{insertiontransformer,levenshteintransformer,diffuser} place edits without committing to positions, but they apply them in a fixed refinement procedure and attach no rate to an edit, so nothing sets the expected number. Edit Flows \citep{editflows2025} and EvoFlows \citep{evoflow} do attach rates and generate in continuous time. Despite the promising aspects of those models, the lack of released codebases and complete training specifications has hindered the adoption of those methods in the field. This paper aims to reconstruct both methods and ground them with the technical details needed for wider adoption.

\begin{figure}[t]
	\begin{center}
		\input{figures/fig-teaser}
	\end{center}
	\caption{\textbf{Balancing protein edit diversity and property conservation of the reference family (i.e., naturalness)}
        We measure diversity as the ratio of mean pairwise Levenshtein distance between the generated set and random real homologs pairs.
		Naturalness is the ratio of homolog-generated pairs correlation and homolog-homolog correlation.
        The real homolog baseline is thus homolog-homolog on both sides of the ratio and produces 1. Evotuning, and evotuning (forced) are simple finetuning methods on the ESM-2 transformer model, and EvoDiff-MSA is a diffusion model. EditJumps is our reformulation of EvoFlows. It defines the optimal frontier between naturalness and diversity. 
		The data is pooled over two heldout antibody families, which motivated the calculation of ratios (see Table~\ref{tab:evoflows-vs}).}
	\label{fig:teaser}
    \vspace{-1.3\baselineskip}
\end{figure}
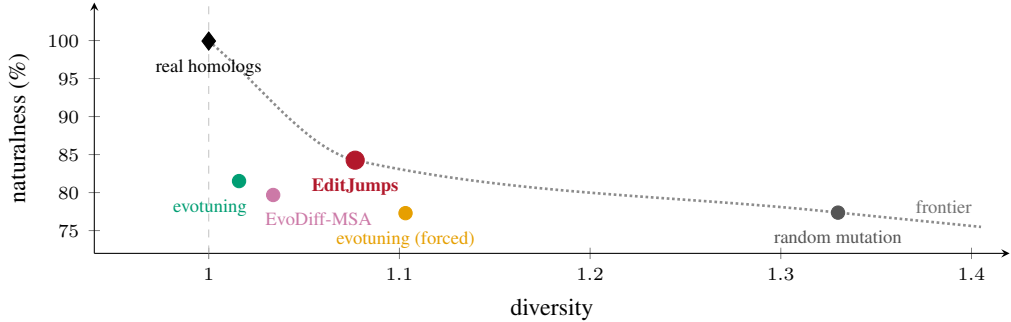

Both originate from flow matching, which regresses a velocity field onto the conditional velocity field of a prescribed probability path \citep{lipman2023flowmatching}. It was formulated on continuous vector spaces, thus the conditions of discrete tokens and variable length don't apply. Later work lifts the first condition by running a continuous-time Markov chain (CTMC) over tokens at fixed length \citep{campbell2024discrete,discreteflowmatching}. Edit Flows lifts the second condition, defining a CTMC on sequences of length at most $N$ with a rate for every single-token insertion, deletion and substitution available in the current state. EvoFlows adapts this to proteins by adding time conditioning and per-position rate and token heads to a pretrained ESM-2 encoder. Edit Flows trains a Llama-architecture transformer \citep{llama3} from scratch, whereas EvoFlows fine-tunes its trunk jointly with the rate heads. We adopt the latter configuration.

While Edit Flows and EvoFlows were originally presented as extensions of flow matching, we re-formalize both methods as the pure-jump case of Generator Matching \citep{generatormatching}, reflecting that discrete sequences evolve through instantaneous jump events rather than continuous flows. This distinction is not merely notational. Because insertion opportunities scale with sequence length, the total jump rate from a state grows unbounded, requiring two necessary safeguards: a non-explosion condition to ensure well-defined waiting times (which Edit Flows achieves via a hard length cap, whereas EvoFlows leaves unconstrained), and a length-dependent clock normalization to prevent the realized jump count from growing with candidate length. Although EvoFlows notes the necessity of clock normalization, it omits both its mathematical formulation and operational value (Section~\ref{sec:background}).

We implement EditJumps, which unifies both methods under a shared pure-jump rate and kernel abstraction, and instantiate the protein editing configuration of EvoFlows. We replicate its training protocol, including source--target sequence coupling, pairwise alignment across unequal lengths, and the induced conditional probability paths. In the absence of a released reference codebase, we evaluate our replication against published derivations, pseudocode, and reported figures, uncovering two primary findings. First, published and replicated edit statistics can be reconciled only by identifying the clock normalization as the parameter governing realized edit count. Second, six of the ten evaluation metrics reported in EvoFlows cannot be deterministically computed from the text alone. Some other metrics like the spectrum Maximum Mean Discrepancy (MMD) estimator is sensitive to reference set size. Our grounded benchmark avoids this confounding normalization (Figure~\ref{fig:teaser}): at matched budgets of $4.1$--$4.6$ edits per sequence, EditJumps produces greater sequence diversity than evotuning \citep{alley2019} and the alignment-based diffusion model EvoDiff-MSA \citep{evodiff} while achieving comparable distributional fidelity, matching unnormalized MMD with standard evotuning on the anti-SARS-CoV-2 nanobody Ty1 and falling within $3\%$ on the anti-HER2 heavy chain HER2-VH. We note that our evaluation is strictly in silico, and the benchmark targets (the camelid single-domain $V_{\text{HH}}$ Ty1 and isolated heavy chain HER2-VH) reflect single-domain constructs, contrasting with the paired heavy-light chain sequences used during generalist pretraining. We make the following \textbf{contributions}: \begin{enumerate*}[label=(\roman*)]
  \item We identify Edit Flows and EvoFlows as instances of the pure-jump case of Generator Matching, with a common jump kernel, and state the non-explosion and clock-normalization conditions a variable-length space requires.

  \item We release \emph{EditJumps}, an open-source PyTorch implementation covering both methods, with checkpoints, configurations and evaluation code for the EvoFlows setting. These are a specified baseline rather than a match to the published values, which the reported conditions do not allow us to recover (Section~\ref{sec:results}).

  \item We train one editor on 1.66M homologous pairs from the paired subset of the Observed Antibody Space \citep{oas} and apply it to held-out targets, removing the per-target retraining the original requires. No per-target-retrained EvoFlows exists to compare against.

  \item We report a replication analysis of EvoFlows: six of ten reported metrics are not deterministically recomputable from the text, replicating the published edit statistics requires an unstated clock normalization, and one metric reverses method rankings under a change of reference set.
\end{enumerate*}

%% file: figures/fig-teaser.tex
\begin{tikzpicture}
\definecolor{cEditJumps}{HTML}{B2182B}
\definecolor{cevotuning}{HTML}{009E73}
\definecolor{cevotuningforced}{HTML}{E69F00}
\definecolor{cEvoDiffMSA}{HTML}{CC79A7}
\definecolor{crandommutation}{HTML}{555555}
\definecolor{crealhomologs}{HTML}{000000}
\begin{axis}[
  width=0.98\linewidth, height=4.9cm,
  xlabel={diversity},
  ylabel={naturalness (\%)},
  ylabel style={align=center}, xmin=0.94, xmax=1.42, ymin=72, ymax=105,
  xtick={1.0,1.1,1.2,1.3,1.4}, ytick={75,80,85,90,95,100},
  tick label style={font=\scriptsize}, label style={font=\small},
  axis lines=left, enlargelimits=false, clip=false,
]
\addplot[gray!45, dashed, forget plot] coordinates {(1,72) (1,105)};
\addplot[black!45, densely dotted, line width=1.0pt, forget plot] coordinates {(1.0000,99.928) (1.0041,99.055) (1.0082,98.127) (1.0124,97.154) (1.0165,96.147) (1.0206,95.117) (1.0247,94.074) (1.0289,93.029) (1.0330,91.993) (1.0371,90.975) (1.0412,89.987) (1.0454,89.040) (1.0495,88.143) (1.0536,87.308) (1.0577,86.544) (1.0619,85.864) (1.0660,85.277) (1.0701,84.793) (1.0742,84.425) (1.0784,84.172) (1.0825,83.949) (1.0866,83.733) (1.0907,83.525) (1.0949,83.324) (1.0990,83.130) (1.1031,82.942) (1.1072,82.761) (1.1114,82.586) (1.1155,82.417) (1.1196,82.253) (1.1237,82.096) (1.1279,81.944) (1.1320,81.797) (1.1361,81.656) (1.1402,81.520) (1.1444,81.388) (1.1485,81.261) (1.1526,81.138) (1.1567,81.019) (1.1609,80.905) (1.1650,80.794) (1.1691,80.687) (1.1732,80.583) (1.1774,80.483) (1.1815,80.386) (1.1856,80.291) (1.1897,80.200) (1.1939,80.111) (1.1980,80.024) (1.2021,79.939) (1.2062,79.856) (1.2104,79.776) (1.2145,79.696) (1.2186,79.618) (1.2227,79.542) (1.2269,79.466) (1.2310,79.391) (1.2351,79.317) (1.2392,79.243) (1.2434,79.170) (1.2475,79.097) (1.2516,79.023) (1.2557,78.950) (1.2599,78.876) (1.2640,78.801) (1.2681,78.725) (1.2722,78.649) (1.2764,78.571) (1.2805,78.492) (1.2846,78.411) (1.2887,78.328) (1.2929,78.244) (1.2970,78.157) (1.3011,78.068) (1.3052,77.977) (1.3094,77.883) (1.3135,77.786) (1.3176,77.686) (1.3217,77.582) (1.3259,77.475) (1.3300,77.365) (1.3331,77.286) (1.3362,77.208) (1.3394,77.130) (1.3425,77.051) (1.3456,76.973) (1.3487,76.894) (1.3519,76.816) (1.3550,76.737) (1.3581,76.659) (1.3612,76.580) (1.3644,76.502) (1.3675,76.423) (1.3706,76.345) (1.3737,76.266) (1.3769,76.188) (1.3800,76.109) (1.3831,76.031) (1.3862,75.952) (1.3894,75.874) (1.3925,75.795) (1.3956,75.717) (1.3987,75.639) (1.4019,75.560) (1.4050,75.482)};
\node[font=\scriptsize, anchor=south east, black!55] at (axis cs:1.4050,76.082) {frontier};
\addplot[only marks, mark=diamond*, mark size=3.4pt, crealhomologs] coordinates {(1.0000,99.93)};
\node[font=\scriptsize, anchor=north, crealhomologs] at (axis cs:1.0000,98.83) {real homologs};
\addplot[only marks, mark=*, mark size=3.4pt, cEditJumps] coordinates {(1.0768,84.26)};
\node[font=\scriptsize\bfseries, anchor=north, cEditJumps] at (axis cs:1.0768,83.16) {EditJumps};
\addplot[only marks, mark=*, mark size=2.5pt, cevotuning] coordinates {(1.0159,81.51)};
\node[font=\scriptsize, anchor=north east, cevotuning] at (axis cs:1.0249,80.41) {evotuning};
\addplot[only marks, mark=*, mark size=2.5pt, cevotuningforced] coordinates {(1.1032,77.29)};
\node[font=\scriptsize, anchor=north, cevotuningforced] at (axis cs:1.1032,76.19) {evotuning (forced)};
\addplot[only marks, mark=*, mark size=2.5pt, cEvoDiffMSA] coordinates {(1.0338,79.68)};
\node[font=\scriptsize, anchor=north west, cEvoDiffMSA] at (axis cs:1.0248,78.58) {EvoDiff-MSA};
\addplot[only marks, mark=*, mark size=2.5pt, crandommutation] coordinates {(1.3300,77.36)};
\node[font=\scriptsize, anchor=north, crandommutation] at (axis cs:1.3300,76.26) {random mutation};
\end{axis}
\end{tikzpicture}

%% file: sections/background.tex
\section{Background}
\label{sec:background}

Although Edit Flows and EvoFlows are formulated in continuous time, the generated sequence itself does not evolve continuously. Instead, it remains unchanged until an insertion, deletion, or substitution moves it to a new sequence. It is the transition rates of these edits that vary with time. This combination of continuous time and discrete changes is characteristic of a jump process \cite{feller1949}
``in a small time interval there is an overwhelming probability that the state will remain unchanged; however, if it changes, the change may be radical''. 
The term also appears in Discrete Walk-Jump Sampling~\citep{dwjs}, where a \emph{jump} is the one-step denoising that projects a noisy sample back onto the data manifold, decoupled from the Langevin walk that precedes it. Here, we use \emph{jump} throughout in the pure-jump sense above: a transition of the
state itself, occurring at a rate.

Generator Matching provides a unified framework for generative modelling through infinitesimal generators, encompassing continuous flows, diffusion processes, and discrete jump processes~\citep{generatormatching}. We use this framework to formalise the sequence-editing dynamics of Edit Flows and EvoFlows as a pure-jump process. We first define the jump process and its generator, then express insertions, deletions, and substitutions as structured jumps. Finally, we establish the conditions under which this formulation holds and introduce its implementation, \emph{EditJumps}.

\subsection{Sequence evolution as a jump process}
We now formalise the sequence-editing dynamics introduced above. Let $x=(x^1,\ldots,x^n)$ denote a finite sequence of length $n$, with each element $x^i$ drawn from a finite vocabulary $\vocab$. We denote by $\XX$ the set of all such sequences. Although $n$ is potentially unbounded, both $\vocab$ and each sequence are finite, making $\XX$ countable.
We consider a continuous-time stochastic process $(x_t)_{t \in [0,1]}$, where $t$ is the time, with $x_0$ drawn from a source distribution (e.g. noise or a start sequence) and $x_1$ from the target data distribution. As described above, $x_t$ remains constant between discrete events, which we call jumps.

Given a point in time $t$ and the current state $x$, the process is specified by a non-negative function $\rate_t(\cdot \mid x_t)$, representing the \emph{rate} at which the state jumps from $x$ to a state $y \neq x$.

We write $\tot_t(x) = \sum_{y \neq x} \rate_t(y \mid x)$ for the total rate of leaving $x$, and $\jump_t(y \mid x) = \rate_t(y \mid x) / \tot_t(x)$ for the probability of the state landing on a new sequence $y$, given that the jump occurs.

\medskip\noindent
The process is then defined by the infinitesimal sampling step:
\begin{equation}
	\label{eq:jump-step}
	x_{t+h} \;=\;
	\begin{cases}
		x_t                            & \text{with probability } 1 - \tot_t(x_t)\,h + o(h),             \\[2pt]
		y \sim \jump_t(\cdot \mid x_t) & \text{with probability } \phantom{1 - {}}\tot_t(x_t)\,h + o(h),
	\end{cases}
\end{equation}
following \citet[Eq.~80]{generatormatching}. For small $h$, the process, therefore, remains in its current state with high probability, while jumps occur with probability proportional to $h$.

Taking the rate of change as $h \to 0$ produces the generator $\gen_t$, which governs the infinitesimal transition from state $x$ at time t. Its action on any test function $f$ is expressed as
\begin{equation}
	\label{eq:generator}
    {(\gen_t f)(x)}
    =
    \underbrace{ 
        \tot_t(x) \sum_{y \neq x} \jump_t(y \mid x)\,\big[f(y) - f(x)\big]
    }_{\substack{\text{total jump rate} \\ \text{$\times$ expected change per jump}}}
	=
    \underbrace{ 
        \sum_{y \neq x} \rate_t(y \mid x)\,\big[f(y) - f(x)\big]
    }_{\text{individual transition rates}}.
\end{equation}

This is the pure-jump case of \citet{generatormatching} (see equations 81-86).

The transition rates $\rate_t(\cdot\mid x)$ that define the target dynamics are generally not available in closed form, therefore the generator $\gen_t$ is parameterised by a neural network. At a given state $x$ and time $t$, the network aims to approximate the \emph{marginal rate} obtained by taking the expected value over the conditional rates of transitioning to x from any other state at time $t$. This can be achieved by optimising a Bregman divergence. A key property of this family of loss functions is that its minimiser is the conditional expectation of the corresponding rate, making it the adequate choice for this task. Depending on data modality, different Bregman divergences can be employed, including squared error for real-valued vectors and cross-entropy for categorical variables.

Once learned, these marginal transition rates fully specify the jump dynamics through~\eqref{eq:generator}. For sampling, they are decomposed into the total jump rate $\tot_t(x)$, which governs the probability of  jumping, followed by a sampling process from the jump kernel $\jump_t(\cdot\mid x)$, which determines its destination. We derive the corresponding sampling procedures in Section~\ref{sec:sampling}.

\subsection{Sequence edits as structured jumps}
Directly parameterising $\rate_t(\cdot \mid x)$ over the sequence space $\XX$ is intractable. Additionally, only states reachable from $x$ through a valid edit have non-zero transition rate. We therefore formulate the transition rates in terms of three distinct single-token edit operations, each acting on an individual element (i.e. amino acid) of the current sequence $x$: $\mathsf{del}_i$ removes $x^i$, $\mathsf{ins}_{i,v}$ inserts $v$ at position $i$, and $\mathsf{sub}_{i,v}$ replaces $x_i$ by $v$, for $1 \le i \le n$ and $v \in \vocab$. Each jump has an explicit structure given by its edit type, position, and, where applicable, the resulting token. We refer to these as \emph{structured jumps}.

Since the target sequence $y$ may be reached through multiple edits $e$, the transition rate between states is obtained by summing the rates of all edits mapping $x$ to $y$:
{}
\begin{equation}
    \label{eq:sum-of-rates}
    \rate_t(y \mid x) = \sum_{e \,:\, e(x) = y} \rate_t(e \mid x).
\end{equation}
{}
Further, we can factorise insertion and substitution rates into a position-specific edit rate and a conditional distribution over the resulting token:
{}
\begin{equation}
	\label{eq:factorisation}
	\rate_t(\mathsf{ins}_{i,v} \mid x) = \rate^{\mathsf{ins}}_t(i \mid x)\;\val^{\mathsf{ins}}_t(v \mid i, x),
	\qquad
	\rate_t(\mathsf{sub}_{i,v} \mid x) = \rate^{\mathsf{sub}}_t(i \mid x)\;\val^{\mathsf{sub}}_t(v \mid i, x),
\end{equation}
{}
with $\val$ a probability distribution over $\vocab$. In contrast, deletion removes the token at position $i$ without introducing a new token and therefore does not require a distribution over $\vocab$. Its rate is given directly by $\rate_t(\mathsf{del}_i \mid x) =\rate^{\mathsf{del}}_t(i \mid x)$.

For every time $t$, sequence $x$, and position $i$, the neural network therefore outputs five quantities: three position-specific rates for insertion, deletion, and substitution, and two token distributions for insertion and substitution. This corresponds to the parameterisation of Edit Flows~\citep[Eqs.~13--15]{editflows2025} and the head structure used in our EvoFlows implementations, \emph{EditJumps}.

\subsection{Non-Explosion and Clock Normalization}
\label{sec:conditions}

On the countable state space $\XX$, the editing dynamics define a continuous-time Markov jump process whose infinitesimal generator is given by~\eqref{eq:generator},
corresponding to the pure-jump formulation of generator matching~\citep{generatormatching}. However, translating this construction to variable-length sequences introduces two distinct regularity challenges that are left unformalized in the original literature.

\paragraph{Non-explosion on unbounded sequence spaces.}
Because insertion operations can occur at any position, the total transition rate $\tot_t(x)$ scales with sequence length $|x_t|$. In CTMC, a pure birth process whose transition rates scale linearly (or superlinearly) with the current state can undergo explosion, accumulating infinitely many jumps and infinite length within a finite time horizon $t \le 1$ \citep{feller1949}.
A jump process is well-posed on $[0,1]$ only if the explosion time $\tau_\infty = \inf\{t : |x_t| = \infty\}$ satisfies $\mathbb{P}(\tau_\infty > 1) = 1$.
In practice, \citet{editflows2025} circumvents explosion by restricting the state space to sequences below a fixed maximum length $L_{\max}$, while \citet{evoflow} leaves the sequence space nominally unbounded.

\paragraph{Length-invariant clock normalization.}
Even when trajectories remain non-explosive, summing transition rates over positions induces an intrinsic length bias: longer sequences accumulate proportionally more edits per unit time than shorter sequences under identical per-position rates.
In therapeutic protein design, lead optimization requires a consistent mutation rate per position regardless of whether the lead is a short peptide or a full variable domain. To decouple the expected edit fraction from sequence length, transition rates must be rescaled by a length-dependent clock factor.
While \citet{evoflow} notes that a clock normalization was applied, the published paper specifies neither the functional form nor the numerical parameter value.
As we demonstrate in Section~\ref{sec:results}, this unstated hyperparameter directly governs the realized edit scale, and tuning it resolves the observed divergence discrepancy between published and reproduced runs.

%% file: sections/setup.tex
\section{Replication Setup}
\label{sec:setup}

Neither Edit Flows \citep{editflows2025} nor EvoFlows \citep{evoflow} released reference code or complete training data specifications.
To reflect the practical demands of therapeutic lead optimization---where retraining per target is computationally prohibitive---we train a single generalist antibody editor across the Observed Antibody Space \citep{oas}, enabling zero-shot editing of novel leads without per-family retraining.
All configurations, data manifests, and model cards will be open-sourced.

\paragraph{Data curation and evaluation targets.}
Training pairs are constructed from paired heavy and light chains ($VH.VL$) in OAS, clustered via MMseqs2 at $0.5$ identity and $0.8$ coverage (see Appendix~\ref{app:data}).
This antibody-wide pretraining restricts downstream evaluation to the two accessible antibody targets from \citet{evoflow}: Ty1 (anti-SARS-CoV-2 VHH) and HER2-VH (anti-HER2 heavy chain), each evaluated across $20$ templates and $200$ held-out homologs.
Non-antibody targets have no homologs in OAS, and anti-EphA2 is unrecoverable in public repositories (Appendix~\ref{app:data}).

\paragraph{Architecture and training protocol.}
We parameterize the editor with an unfrozen, pretrained ESM-2 encoder \citep{esm2}.
Prediction heads receive representations conditioned on sinusoidal time embeddings via Feature-wise Linear Modulation (FiLM): three per-position operation rate heads (insertion, deletion, substitution) and two token prediction heads.
An ESM-2 $35$M-parameter trunk matches the $650$M trunk in realized edit scale ($0.180$ vs.\ $0.182$) and distributional fidelity at a fraction of the computational cost (Appendix~\ref{app:gapcauses}).
The model is fine-tuned for $20{,}000$ steps (batch size $16$, Adam, learning rate $10^{-4}$); Table~\ref{tab:evoflows-vs} reports individual runs and intervals across three seeds (Appendix~\ref{app:runs}).

\paragraph{Supervision discrepancies and deviations.}
\label{sec:gap-sentinel}
Training operates on aligned pairs. When an edit reaches an aligned gap, the published loss and code conflict: Equation~23 correctly assigns zero loss to gap sites, whereas an indexing bug in the reference code mistakenly supervises deletions on adjacent valid residues across 49.2\% of training pairs. (Appendix~\ref{app:gapsentinel}). We adhere to Equation~23.
Four additional operational adaptations depart from the original specifications: linear rather than MLP rate heads (Figure~\ref{fig:arms}), a linear schedule $\kappa_t = t$ matching released code (Appendix~\ref{app:schedule}), frozen holding-interval rates (Appendix~\ref{app:sampling}), and a $0.7$ seed identity threshold for homolog retrieval (Appendix~\ref{app:data}). Further fine-grained deviations and interpretations are listed in the model cards\footnote{\url{https://github.com/VisiumCH/editjumps}}.

\paragraph{The evotuning baseline.}
Following \citet[\S4.2]{evoflow}, evotuning matches the realized mutation budget of EditJumps ($b = 5$ on Ty1, $b = 4$ on HER2-VH).
Target positions are drawn without replacement from a per-column Shannon entropy profile derived from the training alignment (concentrating $50\%$ of mask probability onto just $12\%$ of sequence coordinates) and infilled iteratively with an evotuned ESM-2 MLM.
Because the unforced MLM frequently re-predicts the template residue ($71.7\%$ unchanged on Ty1), we also evaluate a \emph{forced} variant that is not allowed to edit masks back to their value in the previous step, isolating how much of evotuning's fidelity reflects generative editing versus non-intervention. Unlike our generalist editor, evotuning is restricted to substitutions at fixed sequence length and requires retraining a separate 650M trunk for each target family (Appendix~\ref{app:evotune-spec}).

\begin{table}[tbp]
	\caption{EvoFlows' six baselines, reimplemented and evaluated by us (top block) per family: anti-SARS-CoV-2 Ty1, and anti-HER2 construct.
		The two blocks are not comparable; each row compares against its own block's floor and ceiling, respectively random pairing and random mutations.}
	\label{tab:evoflows-vs}
    \vspace{-1\baselineskip}
	\begin{center}
		\resizebox{\textwidth}{!}{%
			\input{tables/table2}}
	\end{center}
    \vspace{-1\baselineskip}
\end{table}

%% file: tables/table2.tex
\begin{tabular}{lrrrrrrrr}
	\toprule
	& \multicolumn{2}{c}{edits/seq} & \multicolumn{2}{c}{diversity} & \multicolumn{2}{c}{MMD $\downarrow$} & \multicolumn{2}{c}{KL $\times 10^{3}$ $\downarrow$} \\
	\cmidrule(lr){2-3}\cmidrule(lr){4-5}\cmidrule(lr){6-7}\cmidrule(lr){8-9}
	method & Ty1 & HER2 & Ty1 & HER2 & Ty1 & HER2 & Ty1 & HER2 \\
	\midrule
	\multicolumn{9}{l}{\emph{ours --- two antibody families}} \\
	random pairing & 23.38 & 23.25 & 23.53 & 23.02 & 0.60 & 0.47 & 0.14 & 0.10 \\
	\textbf{EditJumps} & 4.58 & 4.08 & 24.47 & 24.73 & 0.98 & 1.02 & 0.50 & 0.63 \\
	evotuning & 4.47 & 3.83 & 22.61 & 22.77 & 0.98 & 0.99 & 0.54 & 0.37 \\
	evotuning (forced) & 5.00 & 4.00 & 25.22 & 25.00 & 1.09 & 1.15 & 0.53 & 0.59 \\
	EvoDiff-MSA & 2.96 & 2.61 & 22.85 & 23.23 & 0.97 & 1.11 & 0.61 & 0.42 \\
	random mutations & 4.81 & 3.89 & 30.01 & 29.08 & 1.41 & 1.38 & 0.77 & 0.79 \\
	\addlinespace
	\multicolumn{9}{l}{\emph{the original --- the same two families, its own reported values}} \\
	random pairing & 30.52 & 93.86 & 30.28 & 93.27 & 0.38 & 0.61 & 9.84 & 8.72 \\
	\textbf{EvoFlows} & 14.64 & 35.27 & 35.59 & 90.86 & 1.03 & 1.57 & 21.49 & 23.17 \\
	evotuning & 2.54 & 11.75 & 30.99 & 96.14 & 0.40 & 0.66 & 10.40 & 10.66 \\
	evotuning (forced) & 15.30 & 38.67 & 41.07 & 114.67 & 1.88 & 2.65 & 54.79 & 36.48 \\
	EvoDiff-MSA & 4.43 & 13.80 & 30.01 & 93.06 & 0.42 & 0.66 & 11.63 & 10.16 \\
	random mutations & 13.94 & 35.09 & 48.79 & 133.34 & 2.35 & 3.70 & 80.05 & 86.68 \\
	\bottomrule
\end{tabular}

%% file: sections/results.tex
\section{Results}
\label{sec:results}

Before interpreting our results in Table~\ref{tab:evoflows-vs}, we note that our computed values and published EvoFlows values are not comparable and we could not recover the original experimental conditions from the paper. Against evotuning baselines, EditJumps outperforms forced evotuning across distributional metrics, while demonstrating parity with standard evotuning on $k$-mer MMD despite higher composition divergence and a $\sim 4\%$ increase in edit budget.
Performance across families exhibits domain heterogeneity: MMD differences are $0.08$ on Ty1 and $0.035$ on HER2-VH. On composition KL, EditJumps improves substantially over random mutation on Ty1 while remaining comparable to it on HER2-VH.
All learned models outperform uniform random mutation, yet remain separated from the natural homology floor ($0.98$ MMD versus $0.60$).

\begin{figure}[t]
	\begin{center}
		\input{figures/fig-clock}
	\end{center}
	\caption{\textbf{Sensitivity analysis of the clock normalization hyperparameter.}
		Distributional distance (spectrum MMD) to held-out natural homologs as a function of the clock normalization parameter governing the edit budget, across two antibody families and three random seeds (solid: Ty1; dashed: HER2-VH). For our evaluation we set a scenario where a budget of $\sim 5$ edits was allowed.
        A clock normalization rate that optimizes MMD sits around $80$ or about $8.5$ edits.}
	\label{fig:clock}
\end{figure}
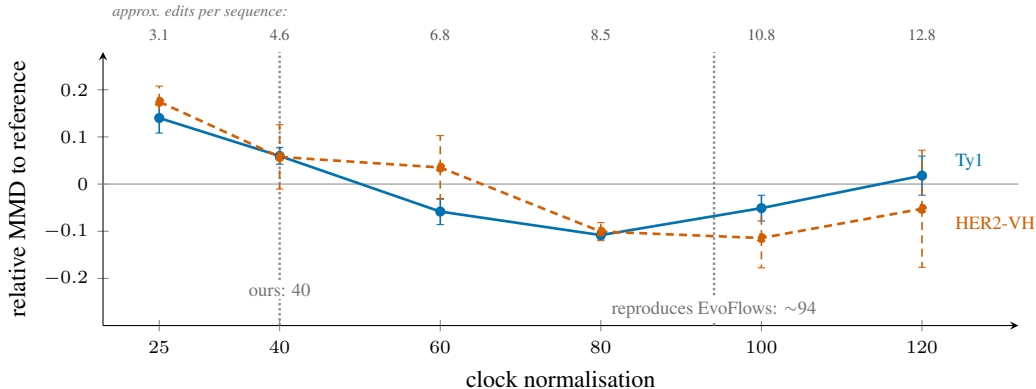

\paragraph{Sensitivity to the clock normalization hyperparameter.}
The edit budget is guided by the clock normalization hyperparameter (\S3.3 of \citet{evoflow}).
A sweep across both families and three seeds reveals that a clock rate of approximately $95$ replicates the published edit ratio of $0.43$ in \cite{evoflow}(Figure~\ref{fig:clock}).

\paragraph{Identifying EvoFlows' 10 evaluation metrics.}
Examining the ten evaluation panels in Figure~3 of \citet{evoflow} reveals that six cannot be deterministically recomputed from the text alone.
Four metrics are well-specified (one modulo an unstated kernel parameter for MMD).
Of the remaining six, three are referenced solely by name but have standard mathematical formalization; two define pairwise matrices (pair covariance and mutual information product) without specifying scalar reductions; and one metric admits multiple incompatible reductions (Levenshtein Distance). Nevertheless, we replicate all 10 metrics in Appendix \ref{app:extended} and in code.

%% file: figures/fig-clock.tex
\begin{tikzpicture}
\definecolor{cty1}{HTML}{0072B2}
\definecolor{cher2}{HTML}{D55E00}
\begin{axis}[
  width=0.98\linewidth, height=5.2cm,
  xlabel={clock normalisation},
  ylabel={relative MMD to reference},
  ylabel style={align=center}, xmin=18, xmax=132, ymin=-0.30, ymax=0.28,
  xtick={25,40,60,80,100,120}, ytick={-0.2,-0.1,0,0.1,0.2},
  yticklabel style={/pgf/number format/fixed, /pgf/number format/precision=2},
  tick label style={font=\scriptsize}, label style={font=\small},
  axis lines=left, enlargelimits=false, clip=false,
]
\addplot[black!35, line width=0.5pt, forget plot] coordinates {(18,0) (132,0)};
\node[font=\tiny, anchor=west, black!55] at (axis cs:18,0.365) {\itshape approx.\ edits per sequence:};
\node[font=\tiny, anchor=south, black!65] at (axis cs:25,0.285) {3.1};
\node[font=\tiny, anchor=south, black!65] at (axis cs:40,0.285) {4.6};
\node[font=\tiny, anchor=south, black!65] at (axis cs:60,0.285) {6.8};
\node[font=\tiny, anchor=south, black!65] at (axis cs:80,0.285) {8.5};
\node[font=\tiny, anchor=south, black!65] at (axis cs:100,0.285) {10.8};
\node[font=\tiny, anchor=south, black!65] at (axis cs:120,0.285) {12.8};
\addplot[black!45, densely dotted, line width=0.9pt, forget plot] coordinates {(40.0,-0.30) (40.0,0.28)};
\node[font=\scriptsize, anchor=south, black!55, fill=white, inner sep=1pt] at (axis cs:40.0,-0.245) {ours: 40};
\addplot[black!45, densely dotted, line width=0.9pt, forget plot] coordinates {(94.1,-0.30) (94.1,0.28)};
\node[font=\scriptsize, anchor=south, black!55, fill=white, inner sep=1pt] at (axis cs:94.1,-0.290) {reproduces EvoFlows: {\raise0.2ex\hbox{$\scriptstyle\sim$}}94};
\addplot[cty1, solid, line width=1.0pt, mark=*, mark size=1.4pt, forget plot,
  error bars/.cd, y dir=both, y explicit,
  error bar style={line width=0.6pt}] coordinates {(25,0.1401) +- (0,0.0319) (40,0.0596) +- (0,0.0175) (60,-0.0583) +- (0,0.0276) (80,-0.1082) +- (0,0.0083) (100,-0.0510) +- (0,0.0273) (120,0.0179) +- (0,0.0415)};
\node[font=\scriptsize, anchor=west, cty1] at (axis cs:123,0.0479) {Ty1};
\addplot[cher2, densely dashed, line width=1.0pt, mark=*, mark size=1.4pt, forget plot,
  error bars/.cd, y dir=both, y explicit,
  error bar style={line width=0.6pt}] coordinates {(25,0.1746) +- (0,0.0331) (40,0.0576) +- (0,0.0683) (60,0.0352) +- (0,0.0676) (80,-0.1008) +- (0,0.0191) (100,-0.1144) +- (0,0.0632) (120,-0.0524) +- (0,0.1242)};
\node[font=\scriptsize, anchor=west, cher2] at (axis cs:123,-0.0824) {HER2-VH};
\end{axis}
\end{tikzpicture}

%% file: sections/discussion.tex
\section{Discussion and Limitations}
\label{sec:discussion}
\label{sec:limitations}

In computational protein engineering, in silico metrics such as sequence recovery, contact-map covariance, and Mutual Information Product (MIP) serve as common surrogates for structural fold preservation and functional viability prior to wet-lab assaying.
However, our findings demonstrate that without grounded empirical baselines, these metrics are easily misinterpreted.
An exact sequence recovery rate of $0.21$ appears modest in isolation, yet represents a $300$-fold increase over chance at matched edit budgets for antibody leads (Appendix~\ref{app:controls}).
Because covariance recovery shifts from $83.6\%$ to $85.1\%$ simply by varying reference partitions, ungrounded metrics risk rewarding sequence conservatism rather than genuine biophysical fitness.

Furthermore, cross-study benchmarking frequently normalises scores.
We show this holds for relative edit distance, but fails for distributional estimators like spectrum MMD: expanding the reference alignment from $200$ to $800$ homologs alters the normalised ratio by $50\%$ on Ty1 and $26\%$ on HER2-VH, a difference sufficiently large to invert method rankings.
Because both the numerator and the baseline floor contract with sample size, internal normalisation cannot replace standardised reference sets.
Encouragingly, core algorithmic formulations remain robust: the symmetric homolog pair enumeration protocol in \citet{evoflow} replicated cleanly without discrepancy (Appendix~\ref{app:held}), showing that replication challenges in discrete protein editing stem primarily from underspecified evaluation protocols rather than algorithmic indeterminacy.

\paragraph{Limitations.}
For antibody lead optimisation, targeted mutations must balance liability removal against paratope disruption.
A key limitation of clock normalisation is the absence of a closed-form mapping to realised edits: because jump rates vary dynamically with time and sequence context, achieving a specific edit budget requires empirical calibration rather than deterministic control.
Biophysically, region-resolved analysis reveals that fidelity is substantially driven by conserved framework scaffolds; within hypervariable CDR loops---which dictate antigen binding---the model closes $<25\%$ of the gap between random mutation and natural homologs (Appendix~\ref{app:region}).
Methodologically, our training regime on concatenated variable domains ($VH.VL$) introduces a domain shift when evaluated on isolated heavy chains and nanobodies across two target families (Ty1 and HER2-VH).
Finally, and exact comparisons against published baseline rows remain constrained by missing intermediate checkpoints in upstream releases (Appendix~\ref{app:release}).

%% file: sections/related.tex
\section{Related Work}
\label{sec:related}

The methods reproduced here combine two lines of generative modelling: discrete generative processes and edit-based sequence generation.

\paragraph{Towards discrete generative processes.}
Diffusion models generate by reversing a fixed corruption of the data \citep{sohldickstein2015diffusion,ddpm}, and structured discrete diffusion carries this construction to discrete token states \citep{d3pm}, as do score-based and continuous-time formulations on those states \citep{seddiscrete,campbell2024discrete}. Flow matching changes the training regime: a conditional probability path from source to target is fixed in advance and regressed onto directly, without simulating a reverse process \citep{lipman2023flowmatching}, and discrete flow matching applies this training to token sequences \citep{discreteflowmatching}. These models hold the sequence at a fixed length and change a position only by substitution. None of them inserts, deletes, or varies the length of a sequence.

\paragraph{Sequence editing.}
Edit-based models generate a sequence through a series of edits, a line of work older than the flow framing. The Insertion Transformer places tokens at arbitrary positions instead of filling a fixed grid of masks \citep{insertiontransformer}. The Levenshtein Transformer adds deletion, so that a model produces the length of an output rather than being given a fixed length \citep{levenshteintransformer}. DiffusER \cite{diffuser} run insertion and deletion as the steps of a denoising process.
These models apply edits through an iterative policy trained against an oracle, in a fixed number of discrete passes, and assign no time or rate to any edit.

\paragraph{Combining discrete generation and edits.}
Edit Flows \citep{editflows2025} and EvoFlows \citep{evoflow} combine the two lines. They take insertion, deletion, and substitution from edit-based generation and assign each a rate under the conditional-path training of flow matching, so that generation runs in continuous time from one sequence to another. The rate makes the expected number of edits a parameter of the sampler, set by the scale of the rates, and makes the generative process a CTMC over sequences.

\paragraph{A jump process.}
Generator matching organises diffusion, flow matching, and pure jump processes as one design space, separated by the form of the generator \citep{generatormatching}. On the countable state space of token sequences the combined construction is a pure jump process. The construction inherits the training path of flow matching and not the dynamics. This motivates our \textit{EditJumps} framing.

\paragraph{Protein generative design, and replication.}
EvoFlows \citep[\S3.4]{evoflow} is framed on protein sequences using a pretrained ESM-2 trunk \citep{esm2}. It sits within a body of generative protein models that produce whole outputs: sequences, whether autoregressive \citep{progen,protgpt2}, antibody infilling \citep{iglm}, or alignment-conditioned \citep{msatransformer,evodiff}, and structure-based backbone design \citep{rfdiffusion,proteinmpnn}, against which the editing formulation outputs a bounded set of residue-level edits to one given sequence.
While surrounding work explores active flow expansion using entropy metrics like the Vendi score \citep{vendi}, our benchmark strictly follows EvoFlows \citep{evoflow} using pooled pairwise distance and $k$-mer MMD. Finally, Discrete Walk-Jump Sampling \citep{dwjs} falls short for antibody lead optimization, because it requires fixing masks to infill, and it cannot calibrate an edit budget (similarly to DiffusER above).

%% file: sections/conclusions.tex
\section{Conclusions}
\label{sec:conclusions}

We formulate the continuous-time sequence editing framework shared by Edit Flows and EvoFlows within a unified formulation of CTMC and pure jump processes.
Framing discrete sequence design as a jump process provides a principled alternative to discrete diffusion and autoregressive generation, equipping discrete token mutations with continuous holding times and operation-specific transition rates (insertion, deletions, substitutions).
For therapeutic lead optimisation, where conservative residue modifications must remediate biophysical liabilities without disrupting antigen binding, this framework establishes a direct bridge between continuous generative flow matching and variable-length sequence engineering.

Our empirical investigation resolves specification challenges in Edit Flows and EvoFlows (Table~\ref{tab:evoflows-vs} and Table~\ref{tab:claims}).
We demonstrated that the apparent divergence in realised sequence novelty between published claims and standard implementations stems from an unstated clock rate normalisation.
Calibrating this edit counts hyperparameter balances sequence diversification against structural preservation (see Figure~\ref{fig:teaser}).
When evaluated against grounded empirical baselines, continuous-time jump models outperform uniform random mutation and forced evotuning, yet remain separated from natural homology floors, highlighting the ongoing challenge of introducing diverse edits while preserving natural sequence context.

Our findings suggest three concrete recommendations for discrete generative biology: (1) parameterizing operational clock rates or adaptive stopping criteria directly within the model, rather than relying on uncalibrated manual rate multipliers; (2) adopting standardized metrics to avoid sample-size artifacts in distributional metrics like MMD; and (3) evaluating generative baselines under both unforced and forced regimes to separate active generative design from passive wild-type conservation.

%% file: sections/ai_statement.tex
\section*{AI Use Statement}

LLM-based assistance was utilized across code development, data analysis, and manuscript preparation under direct author supervision.

\paragraph{Implementation and experiments.}
Training and evaluation pipelines, analysis scripts, and figure-generation routines were developed with LLM-based coding assistance.
All code implementations, algorithmic workflows, and metric definitions were audited, verified, and executed under the direct oversight of the authors.

\paragraph{Writing and analysis.}
LLM assistance was employed to draft and refine portions of the text, format comparison tables, and cross-reference citations against publisher records.
All empirical claims and recovered data points were independently validated against raw artifacts by the authors, who take full intellectual and editorial responsibility for the entire manuscript.

%% file: sections/appendix.tex
\section{Extended results on all metrics}
\label{app:extended}
\label{sec:extended}

In Table \ref{tab:appendix-all}, we report on all 10 metrics in the original EvoFlows methodology. We provide an interpretation of those metrics here and in the code

\input{tables/appendix_table}

\paragraph{Evotuning Baseline Implementation and Sampling Protocol.}
\label{app:evotune-spec}
Following \citet{alley2019} and \citet[\S2.2, \S4.2]{evoflow}, the evotuned
baseline is constructed as a two-stage hybrid:
\begin{enumerate}
    \item \textbf{Family Adaptation (Evotuning):} An \texttt{esm2\_t33\_650M\_UR50D} trunk ($650$M parameters) is fine-tuned independently on the training partition of each antibody family ($36{,}417$ sequences for Ty1; $40{,}489$ for HER2-VH) for $2{,}000$ steps (batch size $16$,
AdamW, learning rate $5 \times 10^{-5}$, $15\%$ masking probability).
    \item \textbf{Positional Entropy Profile (Where to edit):} Multiple
sequence alignments of the family training set define per-column amino-acid
frequencies $p_l(a)$ (excluding gaps and non-standard characters). The positional
sampling distribution is governed by column-wise Shannon entropy:
    \begin{equation}
        H(l) = -\sum_{a \in \mathcal{A}} p_l(a) \log \bigl(p_l(a) +
\epsilon\bigr),
    \end{equation}
    normalized across sequence positions and mapped to template coordinates.
Zero-entropy columns (completely conserved framework residues) are sampled only
after all positive-entropy positions are exhausted.
    \item \textbf{Iterative Infilling (What to edit):} A budget of $b$
positions is sampled from the entropy profile and replaced with \texttt{<mask>}
tokens. The family-adapted MLM infills positions iteratively in random order,
feeding predicted tokens back into context at temperature $T = 1.0$.
    \item \textbf{Budget Matching (Unforced vs.\ Forced):} In the standard
\emph{unforced} regime, the MLM frequently predicts the template residue. To satisfy the requirement of matching the realized mutation count of EditJumps, our implementation tops up masks over up to 8 iterative rounds until $b$ active substitutions are achieved. In the \emph{forced} regime, template residues are masked from the output logits ($-\infty$), guaranteeing $b$ mutations in a
single pass.
\end{enumerate}
Crucially, evotuning is strictly substitution-only and cannot model length variability (indels).

\section{Sampling Algorithms and Discretization Analysis}
\label{app:sampling}
\label{sec:sampling}

\paragraph{Continuous-Time Sampling Formulations.}
Inference in EditJumps simulates trajectories from $t=0$ to $t=1$ under learned transition rates:
\begin{enumerate}
	\item \textbf{Euler $\tau$-leaping:} Partitions the unit interval into $N$ uniform steps of width $h = 1/N$. At each step $t$, the rate field $\rate_t(\cdot \mid x_t)$ is evaluated, and candidate edits fire independently with probability $\rate_t(e \mid x_t) h$. This scheme is first-order accurate in $h$ and computationally efficient, though multiple non-commuting edits may occasionally propose simultaneously within a step.
	\item \textbf{Exact Continuous-Time Simulation (Gillespie):} Simulates event-to-event transitions grid-free. For time-dependent rates, the next jump time $\tau$ satisfies $\int_t^{t+\tau} \tot_s(x_t)\,\mathrm{d}s = -\ln U$, where $U \sim \mathrm{Uniform}(0,1)$. At the event time, the state transitions to $y \sim \jump_{t+\tau}(\cdot \mid x_t)$. In our continuous-time implementation, we freeze rates over the holding interval, yielding an efficient first-order continuous-time sampler.
\end{enumerate}

\paragraph{Empirical Discretization Analysis: Euler vs.\ Exact Gillespie.}
\label{app:euler}
To quantify discretization error introduced by Euler $\tau$-leaping, we benchmarked Euler against exact Gillespie sampling on a fixed synthetic rate field across $2{,}000$ independent trajectories starting from a sequence of length $120$.
Against an exact Gillespie mean of $71.29 \pm 0.15$ edits:
\begin{itemize}
	\item At $N = 2$ steps, Euler over-fires by $4.44 \pm 0.20$ edits;
	\item At $N = 10$ steps, Euler over-fires by $0.76 \pm 0.21$ edits;
	\item At $N = 50$ steps (the setting used throughout all experimental runs), Euler over-fires by $0.08 \pm 0.21$ edits, representing a relative error of $<0.3\%$ and demonstrating that discretization error is negligible.
\end{itemize}
Crucially, clock normalization dictates the direction of discretization error: under clock normalization, expanding sequence length lowers total transition rates, leading Euler's frozen rates to slightly over-fire; in unnormalized regimes, total rates increase with length, causing Euler to under-fire (by $-4.03 \pm 0.42$ edits at $N=2$).
Consequently, sampler discretization does not account for observed empirical discrepancies.

\section{Model Architectures, Training Paths, and Algorithmic Specifications}
\label{app:kernels}

\paragraph{Augmented Training Path Construction.}
Coupling arbitrary source and target sequence pairs $(x_0, x_1)$ is non-trivial due to variable sequence lengths and combinatorial alignment paths.
\citet{editflows2025} and \citet{evoflow} resolve this by lifting sequences to an augmented space $\mathcal{Z} = (\vocab \cup \{\varepsilon\})^N$, where $\varepsilon$ denotes an alignment gap.
Pairs are coupled via optimal Needleman--Wunsch alignment \citep{henikoff1992}.
Each aligned column $(z_0^i, z_1^i)$ specifies a categorical transition:
\begin{itemize}
	\item $\varepsilon \to v$: insertion of token $v \in \vocab$ at position $i$;
	\item $u \to \varepsilon$: deletion of token $u \in \vocab$ at position $i$;
	\item $u \to v$: substitution of token $u$ with $v$.
\end{itemize}
The conditional probability path is defined by coordinate-wise linear mixture:
\begin{equation}
	p_t(z^i \mid z_0^i, z_1^i) \;=\; (1 - \kappa_t)\,\delta_{z_0^i}(z^i) + \kappa_t\,\delta_{z_1^i}(z^i),
\end{equation}
where $\delta$ is the Dirac delta function and $\kappa_t \in [0, 1]$ is a schedule satisfying $\kappa_0 = 0$ and $\kappa_1 = 1$. Thus, at an intermediate time $t$, the target sequence's token at every position is sampled with probability $\kappa_t$ and the source sequence's token with probability $1 - \kappa_t$.

Finally, marginal transition rates are trained by regressing predicted rates onto conditional targets via cross-entropy and Bregman divergence objectives.

\begin{figure}[t]
	\begin{center}
		\input{figures/fig-arms}
	\end{center}
	\caption{\textbf{Head parameterization sensitivity analysis on Ty1.}
		A $2\times2$ factorial comparison of training corpus (OAS vs.\ stock) against head parameterization: the published Appendix~A specification (filled markers; adapting ESM-2 language model heads) versus lightweight linear rate heads with random token heads (open markers).
		Axes are normalized to natural homologs $(1, -1)$.
		Novelty $\Delta$ measures the change in distance relative to template starting coordinates.
		Arrows indicate the effect of altering head parameterization at fixed corpus, shifting novelty by $0.52$ and $1.24$, dominating corpus shifts ($0.15$).}
	\label{fig:arms}
\end{figure}
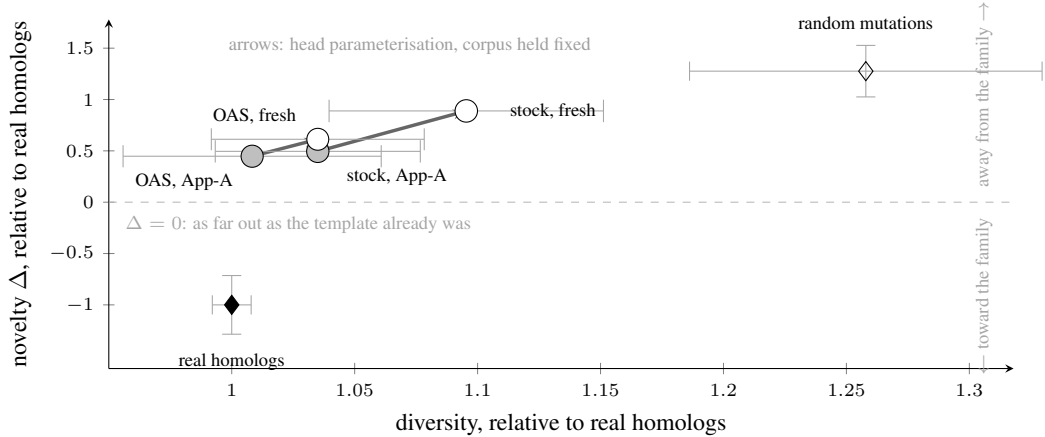

\paragraph{EvoFlows Architecture and Head Parameterization.}
\label{app:evoflows-spec}
\label{app:arms}
EvoFlows utilizes a pretrained ESM-2 trunk fine-tuned jointly \citep{esm2}.
Time conditioning is introduced via sinusoidal embeddings processed through an MLP: $\tau_t = \mathrm{MLP}(\mathrm{Sinusoidal}(t))$, modulating token representations across all heads via shared Feature-wise Linear Modulation (FiLM) scale and shift parameters.
The five output quantities are parameterized as:
\begin{itemize}
	\item Three position-specific rate heads ($\rate^{\mathsf{ins}}, \rate^{\mathsf{del}}, \rate^{\mathsf{sub}}$): parameterized as shallow MLPs mapped to positive values via softplus or bounded sigmoid activations;
	\item Two token prediction heads ($\val^{\mathsf{ins}}, \val^{\mathsf{sub}}$): parameterized either as MLPs from scratch or by adapting ESM-2's pretrained masked language modeling head.
\end{itemize}
Figure~\ref{fig:arms} evaluates the sensitivity of generative dynamics to head parameterization.
Using pretrained ESM-2 language prediction heads yields substantially more conservative edits (novelty $\Delta = 0.45$ vs.\ $0.61$ on OAS; $0.50$ vs.\ $0.89$ on stock), with head parameterization accounting for shifts of $0.52$--$1.24$ in novelty, significantly exceeding the effect of training corpus choice ($0.15$).

\paragraph{Discrepancy in Deletion Supervision in Edit Flows.}
\label{app:gapsentinel}
In \citet[Eq.~23]{editflows2025}, deletion supervision is formally defined for positions where the source token is valid ($a \in \vocab$) and the target is blank ($b = \varepsilon$).
However, inspection of the reference implementation (Figure~13) reveals an off-by-one indexing error.
The code introduces two distinct gap sentinels (\texttt{epsilon\_0\_id} and \texttt{epsilon\_1\_id}) and enters the deletion branch whenever the current token differs from \texttt{epsilon\_0\_id} and the target equals \texttt{epsilon\_1\_id}.
Crucially, the active index into $x_t$ advances only for non-sentinel tokens; when a deletion is encountered, the supervised rate corresponds to the \emph{preceding surviving residue}.
Consequently, the implementation supervises an already deleted position at an incorrect coordinate.
Our reproduction rectifies this discrepancy by strictly following the mathematical formulation of Eq.~23.

\paragraph{Interpolation Schedule Analysis.}
\label{app:schedule}
\citet{editflows2025} reports using a cubic interpolation schedule $\kappa_t = t^3$ in its experimental narrative, yet Figure~13 specifies $\kappa_t = t$ with the code comment \texttt{\# Using a linear schedule}.
Analytically, both schedules deposit identical total cross-entropy mass across training examples ($\int_0^1 \mathrm{d}\kappa_t = 1$).
However, their supervision profiles differ substantially: the linear schedule yields $\mathbb{E}[\kappa_t] = 0.50$, whereas the cubic schedule yields $\mathbb{E}[\kappa_t] = 0.25$, shifting supervision mass toward later times ($t = 0.79$) and increasing the expected count of active edit targets per example from $32.7$ to $50.1$ across $400$ homolog pairs.
We adopted the linear schedule matching the reference code, and confirmed that both schedules remain fully selectable.

\paragraph{Implicit Library Specifications in Sequence Alignment.}
\citet{evoflow} specifies Needleman--Wunsch alignment for training pairs \citep{henikoff1992}, but omits substitution matrix parameters (e.g.\ BLOSUM62) and gap opening/extension penalties.
Because alignment parameters directly determine intermediate corruption targets, training labels are partially governed by third-party library defaults, underscoring that algorithmic specifications in discrete flow models are frequently distributed across external library dependencies.

\section{Dataset Curation, Training Protocol, and Verification}
\label{app:data}

\paragraph{Corpus Curation and Homolog Clustering.}
All training pairs were constructed from the Observed Antibody Space \citep{oas}.
Paired heavy and light chain variable domains ($VH.VL$) were joined by a delimiter token present in the ESM-2 vocabulary.
Homolog clusters were constructed using MMseqs2 in \texttt{easy-cluster} mode with a minimum sequence identity threshold of $0.5$ and coverage threshold of $0.8$.
To prevent high-frequency clonal expansions from dominating training, clusters were capped at $20$ pairs per family, yielding $1{,}660{,}105$ symmetric pairs.
For evaluation targets (Ty1 and trastuzumab HER2-VH), MMseqs2 searches (E-value $\le 0.1$, coverage $\ge 0.8$) were augmented with a $0.7$ seed identity threshold to isolate genuine biological homologs from invariant framework regions.

\paragraph{Algorithmic Verification: Symmetric Pair Enumeration.}
\label{app:held}
\citet[\S4.2]{evoflow} states that training pairs are formed by unordered enumeration of homolog pairs.
We audited this property in our data generation pipeline.
In a random sample of $2{,}000$ training pairs, $44.9\%$ exhibited a higher property score for the second sequence ($44.3\%$ in an independent $4{,}000$-pair audit), consistent with unbiased symmetric enumeration ($\sim 50\%$) and contrasting with oriented pairing regimes ($100\%$).

\paragraph{Training Protocol and Compute Resources.}
\label{app:runs}
Primary model training was executed on a single NVIDIA L4 GPU for $20{,}000$ steps with a batch size of $16$, using the Adam optimizer with a learning rate of $10^{-4}$ ($6\,\mathrm{h}\,29\,\mathrm{m}$ wall-clock time, $1.09\,\mathrm{s}/\mathrm{step}$).
Validation loss dropped rapidly from $798.0$ at initialization to $149.6$ at step $8{,}500$, $138.7$ at step $16{,}000$, and reached $135.2$ at step $19{,}999$.
The model remained non-converged at the $20{,}000$-step budget, with the final four evaluations improving by $2.3$ units over preceding checkpoints.

\paragraph{Operational Integrity and Artifact Verification.}
\label{app:release}
\label{app:silent}
All code, model weights, and evaluation pipelines for EditJumps are released under an MIT license at \url{https://github.com/VisiumCH/editjumps}.
To safeguard experimental reproducibility, our artifact management protocols assert strict integrity constraints:
\begin{itemize}
	\item Training jobs implement synchronous checkpoint validation, verifying that exported model state dictionaries load successfully before reporting job completion.
	\item Metric logging pipelines enforce continuous credential refresh to ensure that loss trajectories and evaluation metrics remain un-truncated across multi-hour runs.
	\item Baseline evaluation scripts verify sequence frame coordinates dynamically, preventing mismatched reference sets between candidate models and natural controls.
\end{itemize}

\section{Evaluation Methodology, Metric Audits, and Ceilings}
\label{app:fig3}

\paragraph{Vector Graphic Extraction and Method Resolution.}
Because \citet{evoflow} reported quantitative results exclusively in Figure~3 without tabulated values, we recovered numerical markers by extracting vector paths from the published PDF and calibrating against axis tick marks.
The six series were resolved unambiguously:
\begin{enumerate}
	\item Rotated $x$-axis tick labels maintain a uniform pitch of $9.87 \pm 0.01\,\mathrm{pt}$, aligning with marker columns at a constant $2.4\,\mathrm{pt}$ offset.
	\item Series ordering (random pairing, EvoFlows, EvoDiff-MSA, evotuning, forced evotuning, random mutation) matches domain endpoints: random pairing achieves optimal quality across all panels, while random mutation establishes the empirical floor.
	\item Figure~5 (page 21) provides the explicit legend omitted in Figure~3, matching marker colors to method labels with zero discrepancy.
\end{enumerate}

\paragraph{Evaluation Metric Completeness Audit.}
Table~\ref{tab:fig3-defined} categorizes the ten metrics reported in \citet{evoflow}.
Only four metrics are completely defined in the text; three are named without mathematical definitions, two provide covariance/MIP matrices without scalar reduction functions, and one leaves the reduction between pooled and per-template averages ambiguous.

\begin{table}[t]
	\caption{\textbf{Completeness of evaluation metrics in \citet{evoflow} Figure~3.}
		Systematic audit of published metric specifications.
		Only four of ten panels are fully reproducible from the text alone (with MMD omitting the kernel hyperparameter $k$).
		Three metrics exist only as panel titles without mathematical definitions or formulas.}
	\label{tab:fig3-defined}
	\begin{center}
		\small
		\begin{tabular}{lll}
			\toprule
			Metric Panel             & Specification Status & Omitted Details / Ambiguities       \\
			\midrule
			Avg Levenshtein to $x_0$ & Defined              & ---                                 \\
			Avg Pairwise Levenshtein & Ambiguous            & Pooled vs.\ within-template average \\
			Covariance Matrix        & Partially defined    & Matrix specified; scalar reduction omitted \\
			ESM-2 PLL                & Defined              & ---                                 \\
			Entropy Delta            & Undefined            & Named in figure only                \\
			Jensen--Shannon Div.     & Undefined            & Named in figure only                \\
			KL Divergence            & Defined              & ---                                 \\
			MIP Matrix               & Partially defined    & Matrix specified; scalar reduction omitted \\
			Spectrum MMD             & Defined              & Kernel parameter $k$ omitted        \\
			Profile Log-Likelihood   & Undefined            & Named in figure only                \\
			\bottomrule
		\end{tabular}
	\end{center}
\end{table}

\paragraph{Sample-Size Dependency and Grounded Baseline Controls.}
\label{app:controls}
Agreement metrics between generated and reference distributions exhibit severe sample-size dependency.
Scoring real homologs against real homologs yields covariance scores of $0.790$ at $N=20$, $0.921$ at $N=100$, and saturates at $0.985$ at $N=800$.
Consequently, raw scores from evaluations with differing reference set sizes are mathematically non-comparable.
Furthermore, evaluation metrics require grounded baselines: an exact sequence recovery rate of $0.21$ appears modest in isolation but represents a $300$-fold enrichment over chance at matched edit budgets; conversely, an apparent agreement score of $0.79$ falls entirely within the random sampling variance of finite sets.

\paragraph{Reference Construction and Overlap Impact.}
\label{app:reference}
Evaluation references ($200$ sequences per family) were drawn from held-out family members.
Because families were mined from OAS, $113$ of Ty1's and $103$ of HER2-VH's reference sequences had been present in the training set.
Re-evaluating with strictly disjoint reference sets shifted spectrum MMD by $-36.7\%$ (Ty1) and $-12.6\%$ (HER2-VH) and increased covariance agreement by $12$--$17$ points, driven by greater sequence homogeneity in the disjoint subset ($23.00$ vs.\ $24.90$ Levenshtein distance).

\paragraph{Ceiling Normalization and Alignment Frame Sensitivity.}
\label{app:ceiling}
\label{app:framecost}
To establish valid cross-study comparisons, metrics must be normalized against attainable ceilings computed on held-out natural homologs (Table~\ref{tab:ceiling}).
However, ceilings are highly sensitive to multiple sequence alignment projections: computing ceilings in an unaligned or mismatched frame shifts MIP agreement by $+2.3$ percentage points, demonstrating that ceilings must be evaluated in identical coordinate frames.

\begin{table}[t]
	\caption{\textbf{Ceiling-normalized performance of EditJumps against EvoFlows.}
		Both models are normalized against their respective natural homolog ceilings.
		EditJumps is evaluated on two antibody families ($200$ references); EvoFlows values are recovered from Figure~3 across six families.
		$^{\dagger}$Jensen--Shannon divergence is evaluated under our positional formulation as the original omits its definition.
		$^{\ddagger}$Pairwise distance is evaluated under the pooled reduction.}
	\label{tab:ceiling}
	\begin{center}
		\small
		\begin{tabular}{lrr}
			\toprule
			Metric                                  & EditJumps (Ours) & EvoFlows (Published) \\
			\midrule
			Covariance (\% of ceiling)              & $84.3$            & $98.4$               \\
			MIP (\% of ceiling)                     & $78.8$            & $95.5$               \\
			Positional JS$^{\dagger}$ ($\times$ floor) & $2.68 \pm 0.10$   & $1.80$               \\
			Pairwise Distance$^{\ddagger}$          & $1.070 \pm 0.011$ & $1.00$               \\
			\bottomrule
		\end{tabular}
	\end{center}
\end{table}

\paragraph{Non-Explanations for the Edit Budget Discrepancy.}
\label{app:gapcauses}
We systematically evaluated alternative hypotheses for the edit budget gap ($0.20$ reproduced vs.\ $0.43$ published):
\begin{itemize}
	\item \textbf{Trunk capacity:} Scaling ESM-2 from $35$M to $650$M shifts the edit ratio from $0.180$ to $0.182$, well within seed variance ($\pm 0.01$).
	\item \textbf{Head parameterization:} Adhering to the published Appendix~A head parameterization is more conservative ($0.19$) than lightweight linear heads ($0.22$--$0.28$).
	\item \textbf{Training duration:} Modulating learning rates at $650$M ($3\times 10^{-4}$ vs.\ $10^{-5}$) yields ratios of $0.17$ and $0.25$, demonstrating that optimization choices move the budget, but clock normalization remains the primary governing mechanism.
\end{itemize}

\begin{figure}[t]
	\begin{center}
		\input{figures/fig-envelope}
	\end{center}
	\caption{\textbf{Evaluation arms within published metric envelopes.}
		Each of the six published series in \citet{evoflow} is plotted as the range spanned across its six evaluation datasets (recovered from Figure~3).
		Our reproduced arms and baseline anchors fall entirely within the published metric envelope.}
	\label{fig:envelope}
\end{figure}
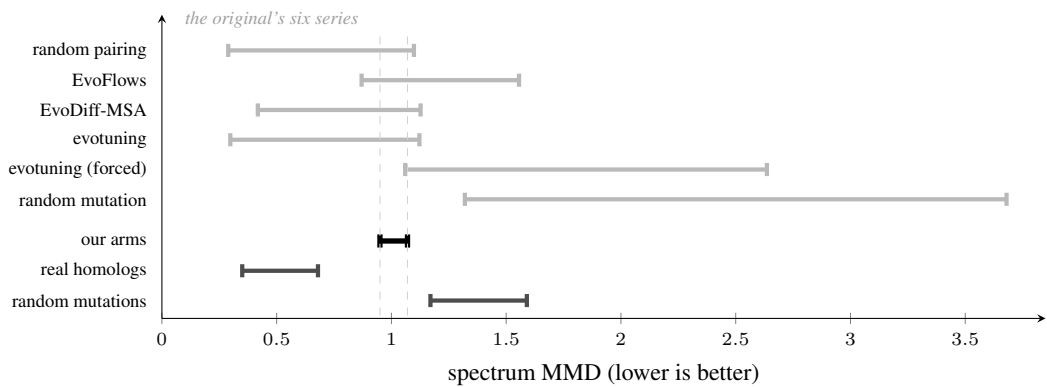

\begin{table}[t]
	\caption{\textbf{Claim-by-claim reproducibility assessment.}
		Operational evaluation of published claims in Edit Flows and EvoFlows.
		Operational completeness denotes whether published texts provide sufficient detail to reproduce reference behavior without external code artifacts.}
	\label{tab:claims}
	\begin{center}
		\small
		\begin{tabular}{ll}
			\toprule
			Published Specification / Claim             & Operational Reproduction Finding           \\
			\midrule
			\multicolumn{2}{l}{\emph{Edit Flows} \citep{editflows2025}}                              \\
			\quad Operational reproducibility from text  & Not answerable (no public code or weights) \\
			\quad Deletion loss supervision             & Discrepancy identified: Eq.~23 vs.\ Fig.~13 code \\
			\midrule
			\multicolumn{2}{l}{\emph{EvoFlows} \citep{evoflow}}                                      \\
			\quad Operational reproducibility from text  & Not answerable (no public code or weights) \\
			\quad Clock normalization hyperparameter    & Omitted in text; governs edit budget \\
			\quad Coevolutionary structure recovery     & Partially recovered ($84.3\%$ / $78.8\%$ of ceiling) \\
			\bottomrule
		\end{tabular}
	\end{center}
\end{table}

\paragraph{Biophysical Analysis: IMGT Regions and Disulfide Retention.}
\label{app:region}
Partitioning antibody sequences via IMGT numbering shows that $76\%$ of residues occupy the structural framework and $24\%$ reside in complementarity-determining regions (CDRs).
In CDRs, where functional sequence diversity is concentrated, EditJumps closes $18.7\%$ (Ty1) and $23.9\%$ (HER2-VH) of the composition KL distance between random mutation and natural homologs.
In framework positions, all methods approach the natural floor ($0.0001$--$0.0011$).
Crucially, EditJumps preserves canonical structural disulfides (IMGT Cys23 and Cys104) in $100\%$ ($393/393$ on Ty1 and $389/389$ on HER2-VH) of numberable generations, compared to $89.0\%$ and $93.8\%$ under random mutation baselines.

%% file: tables/appendix_table.tex
\begin{table}[t]
  \centering\scriptsize
  \setlength{\tabcolsep}{2pt}
  \renewcommand{\arraystretch}{0.95}
  \caption{EvoFlows' \S4.2 reproduced results, on all ten metrics, both seed families, one frame: 20 templates $\times$ 20 variants, holdout 200, agreement ceiling 300 real homologs.}
  \label{tab:appendix-all}
  \begin{tabular}{>{\raggedright\arraybackslash}p{1.9cm} >{\raggedleft\arraybackslash}p{0.80cm} >{\raggedleft\arraybackslash}p{0.86cm} >{\raggedleft\arraybackslash}p{0.80cm} >{\raggedleft\arraybackslash}p{0.78cm} >{\raggedleft\arraybackslash}p{0.85cm} >{\raggedleft\arraybackslash}p{0.80cm} >{\raggedleft\arraybackslash}p{0.80cm} >{\raggedleft\arraybackslash}p{0.80cm} >{\raggedleft\arraybackslash}p{0.80cm} >{\raggedleft\arraybackslash}p{1.05cm}}
    \toprule
    method & edits & pairw.\ Lev. & MMD & KL pos. & KL pool. $\times 10^{3}$ & JS pos. $\times 10^{2}$ & cov. agr. & MIP agr. & $\Delta$H & prof. LL \\
     & -- & $\approx$ & $\downarrow$ & $\downarrow$ & $\downarrow$ & $\downarrow$ & $\uparrow$ & $\uparrow$ & $\to 0$ & $\uparrow$ \\
    \midrule
    \multicolumn{11}{l}{\textit{ty1}} \\
    \mbox{\textbf{EditJumps}} & \mbox{4.58} & \mbox{24.47} & \mbox{0.98} & \mbox{0.07} & \mbox{0.50} & \mbox{1.26} & \mbox{0.91} & \mbox{0.83} & \mbox{0.07} & \mbox{-64.01} \\
    \mbox{rand.\ mut.} & \mbox{4.81} & \mbox{30.01} & \mbox{1.41} & \mbox{0.16} & \mbox{0.77} & \mbox{2.23} & \mbox{0.85} & \mbox{0.66} & \mbox{0.15} & -- \\
    \mbox{rand.\ pair.} & \mbox{23.38} & \mbox{23.53} & \mbox{0.60} & \mbox{0.03} & \mbox{0.14} & \mbox{0.54} & \mbox{0.98} & \mbox{0.94} & \mbox{0.07} & -- \\
    \mbox{EvoDiff} & \mbox{2.96} & \mbox{22.85} & \mbox{0.97} & \mbox{0.07} & \mbox{0.61} & \mbox{1.38} & \mbox{0.89} & \mbox{0.78} & \mbox{0.02} & \mbox{-59.15} \\
    \mbox{Evotune} & \mbox{4.47} & \mbox{22.61} & \mbox{0.98} & \mbox{0.06} & \mbox{0.54} & \mbox{1.14} & \mbox{0.91} & \mbox{0.81} & \mbox{0.01} & \mbox{-57.49} \\
    \mbox{Evotune-f} & \mbox{5.00} & \mbox{25.22} & \mbox{1.09} & \mbox{0.09} & \mbox{0.53} & \mbox{1.40} & \mbox{0.87} & \mbox{0.74} & \mbox{0.09} & \mbox{-67.54} \\
    \mbox{BF-DiT} & \mbox{19.77} & \mbox{33.76} & \mbox{1.58} & \mbox{0.22} & \mbox{0.91} & \mbox{2.17} & \mbox{0.38} & \mbox{0.23} & \mbox{0.25} & \mbox{-95.28} \\
    \mbox{BF-ESM} & \mbox{31.69} & \mbox{44.28} & \mbox{4.31} & \mbox{0.76} & \mbox{10.65} & \mbox{6.28} & \mbox{0.12} & \mbox{0.15} & \mbox{0.33} & \mbox{-155.11} \\
    \midrule
    \multicolumn{11}{l}{\textit{her2vh}} \\
    \mbox{\textbf{EditJumps}} & \mbox{4.08} & \mbox{24.73} & \mbox{1.02} & \mbox{0.05} & \mbox{0.63} & \mbox{1.03} & \mbox{0.93} & \mbox{0.87} & \mbox{0.08} & \mbox{-59.03} \\
    \mbox{rand.\ mut.} & \mbox{3.89} & \mbox{29.08} & \mbox{1.38} & \mbox{0.12} & \mbox{0.79} & \mbox{1.89} & \mbox{0.90} & \mbox{0.78} & \mbox{0.15} & -- \\
    \mbox{rand.\ pair.} & \mbox{23.25} & \mbox{23.02} & \mbox{0.47} & \mbox{0.03} & \mbox{0.10} & \mbox{0.47} & \mbox{0.98} & \mbox{0.94} & \mbox{0.10} & -- \\
    \mbox{EvoDiff} & \mbox{2.61} & \mbox{23.23} & \mbox{1.11} & \mbox{0.06} & \mbox{0.42} & \mbox{1.18} & \mbox{0.91} & \mbox{0.82} & \mbox{0.04} & \mbox{-55.03} \\
    \mbox{Evotune} & \mbox{3.83} & \mbox{22.77} & \mbox{0.99} & \mbox{0.05} & \mbox{0.37} & \mbox{1.02} & \mbox{0.93} & \mbox{0.83} & \mbox{0.03} & \mbox{-52.87} \\
    \mbox{Evotune-f} & \mbox{4.00} & \mbox{25.00} & \mbox{1.15} & \mbox{0.07} & \mbox{0.59} & \mbox{1.21} & \mbox{0.91} & \mbox{0.81} & \mbox{0.10} & \mbox{-62.17} \\
    \mbox{BF-DiT} & \mbox{22.38} & \mbox{37.54} & \mbox{2.09} & \mbox{0.26} & \mbox{1.70} & \mbox{2.67} & \mbox{0.38} & \mbox{0.23} & \mbox{0.35} & \mbox{-104.87} \\
    \mbox{BF-ESM} & \mbox{33.29} & \mbox{45.53} & \mbox{4.69} & \mbox{0.71} & \mbox{14.47} & \mbox{6.30} & \mbox{0.14} & \mbox{0.18} & \mbox{0.40} & \mbox{-153.82} \\
    \bottomrule
  \end{tabular}
\end{table}

%% file: figures/fig-arms.tex
\begin{tikzpicture}
\begin{axis}[
  width=0.97\linewidth, height=6.2cm,
  xlabel={diversity, relative to real homologs},
  ylabel={novelty $\Delta$, relative to real homologs},
  xmin=0.950, xmax=1.318, ymin=-1.62, ymax=1.78,
  xtick={1.0,1.05,1.1,1.15,1.2,1.25,1.3},
  ytick={-1,-0.5,0,0.5,1,1.5},
  tick label style={font=\scriptsize}, label style={font=\small},
  axis lines=left, enlargelimits=false, clip=false,
  error bars/error bar style={gray!70, line width=0.4pt},
]
\addplot[gray!55, dashed, forget plot] coordinates {(0.950,0) (1.318,0)};
\node[anchor=north west, font=\scriptsize, gray!70]
  at (axis cs:0.953,-0.05) {$\Delta=0$: as far out as the template already was};
\node[anchor=west, font=\scriptsize, gray!70, rotate=90]
  at (axis cs:1.306,0.06) {away from the family $\rightarrow$};
\node[anchor=east, font=\scriptsize, gray!70, rotate=90]
  at (axis cs:1.306,-0.06) {$\leftarrow$ toward the family};
\addplot[only marks, mark=diamond*, mark size=3.6pt, black,
  error bars/.cd, x dir=both, x explicit, y dir=both, y explicit]
  coordinates {(1.0000,-1.0000) +- (0.0079,0.2857)};
\node[font=\scriptsize, anchor=north] at (axis cs:1.0000,-1.3757) {real homologs};
\addplot[only marks, mark=diamond, mark size=3.6pt, black,
  error bars/.cd, x dir=both, x explicit, y dir=both, y explicit]
  coordinates {(1.2579,1.2762) +- (0.0717,0.2508)};
\node[font=\scriptsize, anchor=south] at (axis cs:1.2579,1.5970) {random mutations};
\addplot[-{Stealth[length=5pt]}, black!60, line width=1.3pt, forget plot]
  coordinates {(1.0083,0.4476) (1.0350,0.6127)};
\addplot[-{Stealth[length=5pt]}, black!60, line width=1.3pt, forget plot]
  coordinates {(1.0350,0.4952) (1.0954,0.8889)};
\node[anchor=west, font=\scriptsize, gray!75] at (axis cs:0.995,1.52)
  {arrows: head parameterisation, corpus held fixed};
\addplot[only marks, mark=*, mark size=4.16pt, draw=black, fill=black!25,
  error bars/.cd, x dir=both, x explicit, y dir=both, y explicit]
  coordinates {(1.0083,0.4476) +- (0.0525,0.0222)};
\node[font=\scriptsize, anchor=north east] at (axis cs:1.0043,0.3726) {OAS, App-A};
\addplot[only marks, mark=*, mark size=4.20pt, draw=black, fill=black!25,
  error bars/.cd, x dir=both, x explicit, y dir=both, y explicit]
  coordinates {(1.0350,0.4952) +- (0.0417,0.0508)};
\node[font=\scriptsize, anchor=north west] at (axis cs:1.0430,0.4202) {stock, App-A};
\addplot[only marks, mark=*, mark size=4.12pt, draw=black, fill=white,
  error bars/.cd, x dir=both, x explicit, y dir=both, y explicit]
  coordinates {(1.0350,0.6127) +- (0.0433,0.0190)};
\node[font=\scriptsize, anchor=south east] at (axis cs:1.0300,0.6727) {OAS, fresh};
\addplot[only marks, mark=*, mark size=4.22pt, draw=black, fill=white,
  error bars/.cd, x dir=both, x explicit, y dir=both, y explicit]
  coordinates {(1.0954,0.8889) +- (0.0558,0.0159)};
\node[font=\scriptsize, anchor=west] at (axis cs:1.1094,0.8889) {stock, fresh};
\end{axis}
\end{tikzpicture}

%% file: figures/fig-envelope.tex
\begin{tikzpicture}
\begin{axis}[
  width=0.95\linewidth, height=5.6cm,
  xlabel={spectrum MMD (lower is better)},
  xmin=0, xmax=3.85, ymin=0, ymax=10.2,
  xtick={0,0.5,1,1.5,2,2.5,3,3.5},
  ytick={9,8,7,6,5,4,2.6,1.6,0.6},
  yticklabels={{random pairing},{EvoFlows},{EvoDiff-MSA},{evotuning},{evotuning (forced)},{random mutation},{our arms},{real homologs},{random mutations}},
  tick label style={font=\scriptsize}, label style={font=\small},
  axis lines=left, enlargelimits=false, clip=false, ytick style={draw=none},
]
\node[anchor=west, font=\scriptsize\itshape, gray!75]
  at (axis cs:0.06,10.05) {the original's six series};
\addplot[gray!55, line width=1.6pt, forget plot] coordinates {(0.289,9.00) (1.098,9.00)};
\addplot[gray!55, line width=1.6pt, forget plot] coordinates {(0.289,8.78) (0.289,9.22)};
\addplot[gray!55, line width=1.6pt, forget plot] coordinates {(1.098,8.78) (1.098,9.22)};
\addplot[gray!55, line width=1.6pt, forget plot] coordinates {(0.870,8.00) (1.556,8.00)};
\addplot[gray!55, line width=1.6pt, forget plot] coordinates {(0.870,7.78) (0.870,8.22)};
\addplot[gray!55, line width=1.6pt, forget plot] coordinates {(1.556,7.78) (1.556,8.22)};
\addplot[gray!55, line width=1.6pt, forget plot] coordinates {(0.418,7.00) (1.127,7.00)};
\addplot[gray!55, line width=1.6pt, forget plot] coordinates {(0.418,6.78) (0.418,7.22)};
\addplot[gray!55, line width=1.6pt, forget plot] coordinates {(1.127,6.78) (1.127,7.22)};
\addplot[gray!55, line width=1.6pt, forget plot] coordinates {(0.298,6.00) (1.122,6.00)};
\addplot[gray!55, line width=1.6pt, forget plot] coordinates {(0.298,5.78) (0.298,6.22)};
\addplot[gray!55, line width=1.6pt, forget plot] coordinates {(1.122,5.78) (1.122,6.22)};
\addplot[gray!55, line width=1.6pt, forget plot] coordinates {(1.060,5.00) (2.636,5.00)};
\addplot[gray!55, line width=1.6pt, forget plot] coordinates {(1.060,4.78) (1.060,5.22)};
\addplot[gray!55, line width=1.6pt, forget plot] coordinates {(2.636,4.78) (2.636,5.22)};
\addplot[gray!55, line width=1.6pt, forget plot] coordinates {(1.320,4.00) (3.680,4.00)};
\addplot[gray!55, line width=1.6pt, forget plot] coordinates {(1.320,3.78) (1.320,4.22)};
\addplot[gray!55, line width=1.6pt, forget plot] coordinates {(3.680,3.78) (3.680,4.22)};
\addplot[black, line width=2.0pt, forget plot] coordinates {(0.950,2.60) (1.070,2.60)};
\addplot[black, line width=2.0pt, forget plot] coordinates {(0.950,2.38) (0.950,2.82)};
\addplot[black, line width=2.0pt, forget plot] coordinates {(1.070,2.38) (1.070,2.82)};
\addplot[black!70, line width=1.6pt, forget plot] coordinates {(0.350,1.60) (0.680,1.60)};
\addplot[black!70, line width=1.6pt, forget plot] coordinates {(0.350,1.38) (0.350,1.82)};
\addplot[black!70, line width=1.6pt, forget plot] coordinates {(0.680,1.38) (0.680,1.82)};
\addplot[black!70, line width=1.6pt, forget plot] coordinates {(1.170,0.60) (1.590,0.60)};
\addplot[black!70, line width=1.6pt, forget plot] coordinates {(1.170,0.38) (1.170,0.82)};
\addplot[black!70, line width=1.6pt, forget plot] coordinates {(1.590,0.38) (1.590,0.82)};
\addplot[gray!35, dashed, forget plot] coordinates {(0.950,0.1) (0.950,9.6)};
\addplot[gray!35, dashed, forget plot] coordinates {(1.070,0.1) (1.070,9.6)};
\end{axis}
\end{tikzpicture}